\documentclass[letterpaper, 10 pt, conference]{ieeeconf}  

\IEEEoverridecommandlockouts                              

\usepackage{float}
\usepackage{xcolor}
\usepackage{graphicx}
\usepackage{tabularx}
\usepackage{booktabs}
\usepackage{makecell}
\usepackage{subcaption}
\usepackage{pifont}
\usepackage{placeins}
\usepackage{url}

\newcolumntype{Y}{>{\centering\arraybackslash}X}

\title{\LARGE \bf
SORT3D: Spatial Object-centric Reasoning Toolbox for Zero-Shot 3D Grounding Using Large Language Models
}

\author{Nader Zantout$^{*,1}$ \and 
Haochen Zhang$^{*,1}$ \and
Pujith Kachana$^{1}$ \and
Jinkai Qiu$^{1}$ \and
Guofei Chen$^{1}$ \and
Ji Zhang$^{1}$ \and
Wenshan Wang$^{1}$ 
\thanks{$^{*}$ Denotes equal contribution.}
\thanks{$^{1}$The authors are with Carnegie Mellon University, Robotics Institute, Pittsburgh, PA.
{\tt\small\{nzantout, haochen4, pkachana, jinkaiq, guofeic, zhangji, wenshanw\}@andrew.cmu.edu}}
}

\begin{document}

\maketitle
\thispagestyle{empty}
\pagestyle{empty}

\begin{abstract}
Interpreting object-referential language and grounding objects in 3D with spatial relations and attributes is essential for robots operating alongside humans. However, this task is often challenging due to the diversity of scenes, large number of fine-grained objects, and complex free-form nature of language references. Furthermore, in the 3D domain, obtaining large amounts of natural language training data is difficult. Thus, it is important for methods to learn from little data and zero-shot generalize to new environments. To address these challenges, we propose SORT3D, an approach that utilizes rich object attributes from 2D data and merges a heuristics-based spatial reasoning toolbox with the ability of large language models (LLMs) to perform sequential reasoning. Importantly, our method does not require text-to-3D data for training and can be applied zero-shot to unseen environments. We show that SORT3D achieves state-of-the-art zero-shot performance on complex view-dependent grounding tasks on two benchmarks. We also implement the pipeline to run real-time on two autonomous vehicles and demonstrate that our approach can be used for object-goal navigation on previously unseen real-world environments. All source code for the system pipeline is publicly released. \footnote{https://github.com/nzantout/SORT3D}
\end{abstract}

\section{Introduction}
\begin{figure}[htbp]
    \centering
    \includegraphics[width=0.85\linewidth, height=0.8\linewidth]{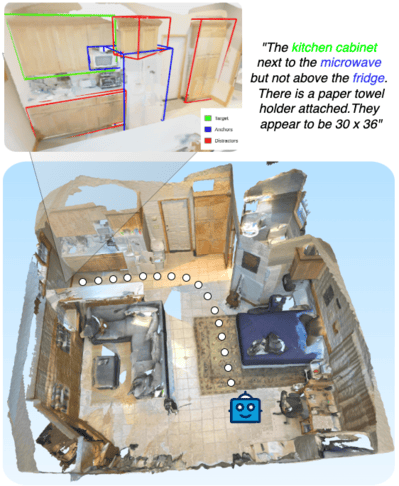}
    \setlength{\belowcaptionskip}{-8pt}
    \caption{An example of our system's workflow for using referential object grounding for downstream object-goal navigation. The agent uses the 2D image for fine-grained grounding in the presence of distractor objects.}
    \label{fig:example}
\end{figure}

As we progress toward generalizable robots operating in human-centered environments like homes and offices, it is crucial for these agents to interact through natural language and align visual observations with natural language references. This capability is essential for applications such as robot caregivers and indoor assistants. Resolving natural language expressions referring to specific objects using semantic object attributes and inter-object spatial relations---the core challenge of \textbf{3D referential grounding}---remains difficult despite being an intuitive task for humans. For example, understanding statements such as ``\textit{the chair closest to the closet door}", is a task trivial for humans \cite{achlioptas2020referit3d} but still challenging for robots. While humans are usually able to identify objects from referring expressions by filtering out irrelevant objects, reasoning about spatial relationships, and utilizing semantic object attributes, such tasks remain challenging for state-of-the-art (SOTA) methods due to several reasons. First, indoor environments often contain numerous objects belonging to fine-grained classes \cite{ramakrishnan2021habitat}, with distributions that vary widely across different homes and environments. Second, training end-to-end learning-based methods on 3D referential grounding requires a large amount of annotated data aligning language references to a 3D scene, which the 3D domain lacks in comparison to the 2D vision-language domain \cite{achlioptas2020referit3d}.

While a number of existing works have developed end-to-end methods to tackle this task through training multi-modal alignment with large transformer models \cite{jain2022bottom, huang2022multi, zhu20233d}, these methods require large-scale annotated data, often overfitting to specific syntactic structure in training datasets while struggling to generalize to more complex utterances, resulting in mediocre performance. More recently, numerous works have leveraged the reasoning capabilities and rich language semantics of large language models (LLMs) for 3D referential grounding \cite{fang2024transcrib3d, hsu2023ns3d, chen2024grounded, yuan2024solving, yang2024llm, xu2024vlm, li2024seeground}. While some achieve strong results on benchmark datasets \cite{fang2024transcrib3d}, complex real-world natural language expressions that employ both semantic attributes and spatial relations such as ``the tall recycling bin to the left if you are facing the door” remain challenging. Many LLM-based approaches either struggle with poor zero-shot performance or rely on careful fine-tuning or heavy prompt engineering tailored towards benchmark datasets. Consequently, such methods are not typically designed for system deployment, relying on high fidelity reconstructed meshes of 3D scenes \cite{dai2017scannet, Matterport3D} and failing to account for constraints in model size, efficiency, and noise in the real-time semantic mapping process.

To this end, we propose \textbf{SORT3D}, a \textbf{S}patial \textbf{O}bject-centric \textbf{R}easoning \textbf{T}oolbox for \textbf{3D} Grounding Using LLMs, shown in Figure \ref{fig:pipeline}, as a novel pipeline for 3D spatial reasoning tailored towards the downstream application of object-goal navigation \cite{batra2020objectnav}. SORT3D is modeled intuitively after human reasoning in object disambiguation, enabling data-efficient 3D referential grounding without requiring annotated 3D training data. To achieve this, we decompose the task into a three-stage approach. After obtaining object names and bounding boxes using an instance-level semantic mapping module, we leverage SOTA 2D vision-language models (VLMs) to extract semantic object attributes important for distinguishing objects. We then use an open-vocabulary object filtering module to efficiently filter relevant objects in large scenes with hundreds of objects, similar to how humans focus their attention to relevant objects. We finally leverage the strong semantic language and sequential reasoning priors of LLMs using chain-of-thought prompting \cite{wei2022chain} to parse a complex referential statement into a series of function calls to our \textbf{spatial reasoning toolbox}, containing search functions that return the IDs of objects that satisfy heuristically defined elementary spatial relations. This delegates complex relative spatial reasoning to structured rule-based logic, ensuring greater accuracy and interpretability. As a result, our method only requires a single in-context example of the toolbox usage and no other training data.

We evaluate our method on standard 3D object referential grounding benchmarks, ReferIt3D \cite{achlioptas2020referit3d} and IRef-VLA \cite{zhang2025irefvlabenchmarkinteractivereferential}, and demonstrate performance competitive with SOTA on complex view-dependent statements while requiring less data to zero-shot generalize. We also deploy our full pipeline on two robotic ground vehicles for real-time indoor navigation, demonstrating our method's ability to further generalize to previously unseen environments and run online alongside real-time perception and autonomy stacks.

\section{Related Work}
\subsection*{Referential Object Grounding Datasets}
Using referring expressions to identify a target object in a 3D scene, defined as the 3D referential grounding task, has been explored in a number of datasets such as ReferIt3D \cite{achlioptas2020referit3d}, ScanRefer \cite{chen2020scanrefer}, SceneVerse \cite{jia2024sceneverse}, and IRef-VLA \cite{zhang2025irefvlabenchmarkinteractivereferential}. Of these datasets, only the Nr3D subset of ReferIt3D and ScanRefer contain human-generated natural language utterances, while the remaining datasets generate synthetic statements using template-based generation and/or LLMs to rephrase. Synthetic datasets only employ basic spatial relations like "lamp by the desk" in referential statements, although IRef-VLA adds the usage basic semantic object attributes---color and size---to create unambiguous references. Compared to the template-generated datasets, the human-generated statements in Nr3D contain far more complex spatial grounding language, such as ``Facing the three boxes, it is the box on the right that is on top of the larger blue box.", which requires view-point grounding, ``The lamp to the left of the desk. NOT the lamp between the beds", which contains negation, and ``the chair closest to the metal appliance" which uses coarse object references and semantic attribute. This complexity, coupled with the small scale of human-generated 3D referential language data, motivates the need for 3D referential grounding methods to leverage strong semantic language priors from other sources (like LLMs), and use a more structured approach for grounding.

\subsection*{3D Referential Object Grounding Baselines}

Grounding object references to 3D scenes has been explored more extensively by various methods since the introduction of benchmarks specific to the task. Approaches to this task include end-to-end models like BUTD-DETR \cite{jain2022bottom}, MVT \cite{huang2022multi}, ViL3DRel \cite{chen2022language}, 3D-VisTA \cite{zhu20233d}, and GPS trained on SceneVerse \cite{jia2024sceneverse}, which fuse multi-modal information in large transformer models and are trained and fine-tuned directly on referential grounding benchmarks. More recent methods decompose the task, leveraging neuro-symbolic frameworks like NS3D \cite{hsu2023ns3d}, and LLMs and VLMs in a zero-shot manner to effectively reason about spatial relations like ZSVG3D \cite{yuan2024visual}, VLM-Grounder \cite{xu2024vlm}, CSVG \cite{yuan2024solving}, and Transcrib3D \cite{fang2024transcrib3d}. Utilizing the reasoning capabilities of LLMs in the text domain, Transcrib3D achieves overall accuracies of 70.2\% and 98.4\% respectively on subsets of Nr3D and Sr3D, outperforming all previous methods. To achieve this, Transcrib3D represents a scene as a list of object names, colors, and positions, and relies on iterative code generation and benchmark-specific prompt engineering giving the LLM guiding principles for grounding. Transcrib3D additionally fine-tunes smaller LLMs on incorrect answers with corrections self-reasoned by larger LLMs. While significant progress in grounding accuracy has been made, the complex spatial reasoning required for the Nr3D dataset, especially with statements that involve egocentric viewpoints and utilize object semantics, continues to pose a challenge, especially for zero-shot methods \cite{xu2024vlm, yuan2024visual, yuan2024solving, fang2024transcrib3d}. We aim to address this gap with our method by combining structured heuristics and vision foundation models with the sequential reasoning capabilities of LLMs.

\subsection*{Grounding Large Language Models in 3D}
Recent efforts have also aimed to develop 3D foundation models capable of handling general 3D tasks. These models leverage the strong reasoning capabilities of pretrained LLMs, achieving strong performance on 3D scene understanding tasks by fine-tuning the LLMs on tokenized 3D data. One of the pioneering works in this space, 3D-LLM \cite{3dllm}, distills rich 2D foundation model features and a 2D VLM backbone to improve performance, mitigating the lack of 3D data. \cite{chen2024grounded} improves upon the work to handle grounding to different viewpoints and generalizes to various 3D tasks with large-scale language-scene pretraining. The efficacy of 2D foundation model features in these approaches demonstrates that grounding 3D understanding in well-established 2D and language-based representations is a powerful approach for advancing 3D reasoning. This factor subsequently inspires the design of our pipeline with the incorporation of 2D captions.

\subsection*{Vision-and-Language Navigation with LLMs}
A number of recent works have also leveraged the sequential reasoning capabilities of LLMs for navigation tasks. NavGPT \cite{zhou2024navgpt} and NavCoT \cite{lin2024navcot} use VLMs to generate text descriptions of viewpoints in the scene in 2D, then task the LLM with selecting the next action in an instruction-following task. Such methods demonstrate the effectiveness of leveraging 2D visual information to guide grounding and action selection in the 3D space. However, within a collaborative setting between a human and a robotic agent, the human would more commonly use single referential statements like "fetch the tv remote on the cabinet" and assume the robot has full knowledge the scene rather than describing a full trajectory towards an object. For building a 3D referential grounding-based VLN system with object-goal navigation as the basic downstream task, text descriptions of landmarks must also be combined with a structured map representing the scene.


\section{Methodology}
\begin{figure*}
    \centering
    \includegraphics[width=0.85\textwidth, height=0.34\textwidth]{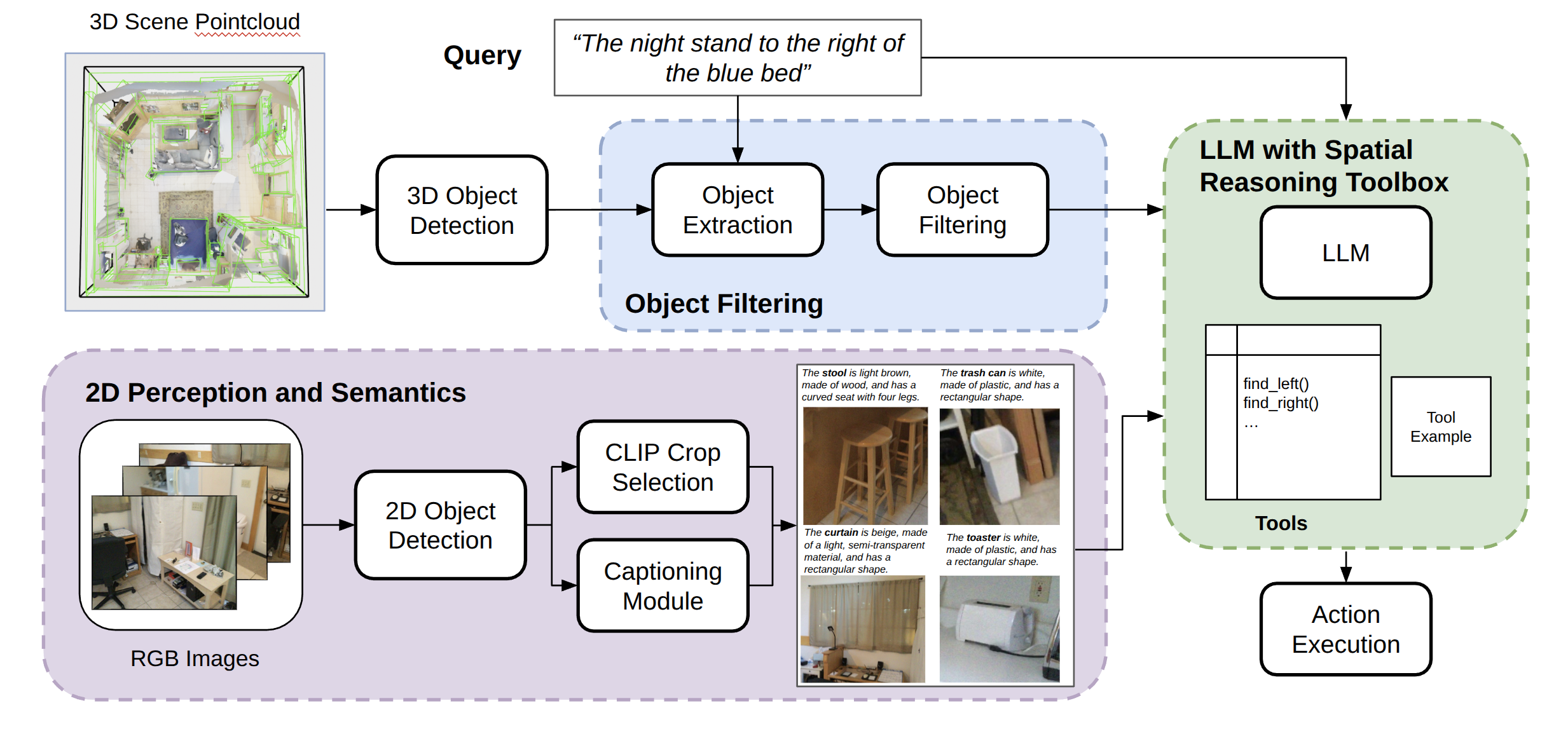}
    \caption{The full system diagram for the SORT3D framework}
    \label{fig:pipeline}
\end{figure*}

In a human-agent collaborative setting, humans commonly refer to objects in the scene by relating them to other objects using commands like "grab the red mug in the top left cupboard over the sink". Finding the correct object being referenced from the utterance is the task of \textbf{3D referential grounding}, which is fundamental for the deployment of a practical VLN system. 3D referential grounding additionally acts as a precursor to downstream tasks such as object-goal navigation, multi-action instruction-following, and scene visual question answering (VQA). We therefore present \textbf{SORT3D}, a zero-shot pipeline for 3D referential grounding, which decomposes the task into multiple subtasks, leverages foundation models to obtain robust zero-shot performance, and targets downstream mobile robot navigation in a real-world environment.

The input to the grounding pipeline consists of perception information from the scene and a free-form referring expression in natural language. The output is the ID of the target object referenced. Figure \ref{fig:pipeline} shows our proposed framework, which can be broken down into four components: (i) an instance-level semantic mapping system to obtain 3D bounding boxes for real-world deployment, (ii) a captioning pipeline to incorporate rich 2D semantic information for each object, (iii) filtering for relevant objects based on the input utterance, and (iv) LLM-based reasoning augmented with a spatial reasoning toolbox to resolve the target object, followed by code generation for executing a downstream action. Each of these components is described in further detail below.

\subsection{Instance-level Semantic Mapping}

For our real-world experiments, we use an object instance-level semantic mapping module running in real-time to obtain the 3D bounding boxes to be input into the LLM and the spatial reasoning toolbox. This component is the only pipeine change required for our method to be deployed in the real-world. Our mobile robot perception setup for real-world experiments consists of a 360 camera and a 3D LiDAR (section \ref{sec:real-world} contains further hardware details). We initially perform object detection in 2D using open-vocabulary object detection \cite{liu2024groundingdinomarryingdino} and instance segmentation \cite{ravi2024sam2segmentimages} models. We then project the registered LiDAR point clouds onto the semantic images and associate each point with its corresponding pixel semantic ID. As the robot moves and produces new observations, we associate per-frame object instance pointclouds using a 2D tracking module and 3D proximity priors, followed by filtering steps to obtain 3D instance pointclouds\footnote{For further details on the semantic mapping module, see https://github.com/gfchen01/semantic\_mapping\_with\_360\_camera\_and\_3d\_lidar}. The usage of open-vocabulary 2D foundation models allows our semantic mapping module to generalize to new environments as we show in our real-world experiments (section \ref{sec:real-world}). For our results on the ReferIt3D benchmarks, we simply use ground truth bounding boxes and instance segmentations. 

\subsection{Enhancing Object Perception with 2D Captions}

Accurately understanding the attributes and affordances of 3D objects is an essential first step for referential grounding. Existing works \cite{fang2024transcrib3d}, \cite{huang2022multi} use high-level information such as object bounding boxes and labels for this task, often acquired through 3D segmentation. While 3D segmentation provides useful object-centric information, it often fails to capture fine-grained attributes like color and shape, which is a key bottleneck of 3D object grounding models \cite{zhu20233d}. Nonetheless, large-scale training for accurate 3D perception is still a challenging problem due to the lack of data and task complexity. On the other hand, VLMs trained on vast amounts of 2D data have strong priors for object understanding in 2D. Notably, VQA models excel at captioning objects in a scene. Therefore, we leverage 2D VQA models to generate descriptions for 3D objects in the scene, providing richer visual details for grounding that are otherwise missed with pure 3D perception. This approach also mirrors how humans perceive and identify objects: narrowing down candidate objects through attributes and relations to resolve ambiguities. In our approach, this finer-grained inspection is achieved by generating captions for each object from cropped object images, providing a more intuitive and precise form of object grounding.

\begin{figure}[htbp]
    \centering
    \includegraphics[width=0.34\textwidth]{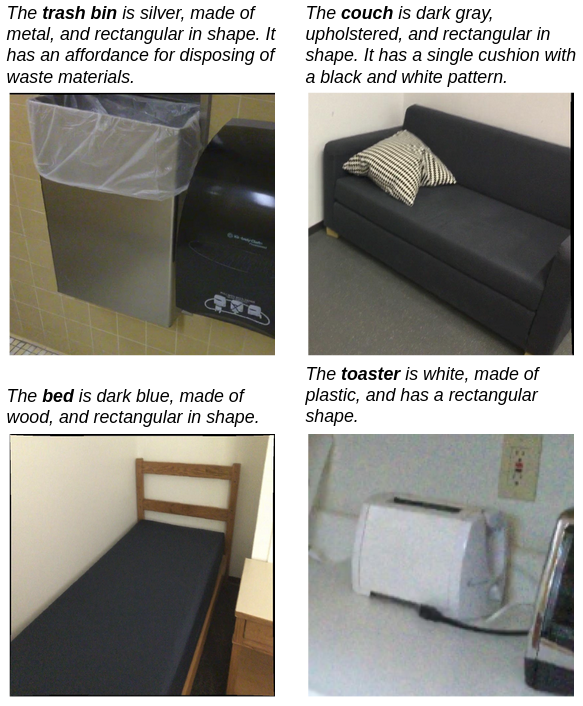}
    \caption{Generated image crops and corresponding caption}
    \label{fig:captions}
\end{figure}

A key decision to be made for captioning is what image to give the VQA model when multiple views of an object are present. We make this choice by using the viewpoint that has the \textit{highest CLIP similarity} with the target label, which is obtained from ground truth for the ReferIt3D benchmarks and from the semantic mapping module for the real-world experiments. We use Qwen2-VL-7B \cite{yang2024qwen2technicalreport} as our VLM as we found it to perform best in generating accurate and concise descriptions following our template, and the quantized version of Qwen2.5-VL-Instruct-3B for system deployment due to memory constraints. We query the VLM with the following prompt format: \texttt{``You are an AI model that describes the characteristics of a query object in an image. Describe the $<$object$>$ in this image, using properties like color, material, shape, affordances, and other meaningful attributes. Provide the response in this format: `The $<$object name$>$ is $<$color$>$, $<$material$>$, $<$shape$>$'"}. We release all object crops and captions as a supplement to the ScanNet \cite{dai2017scannet} dataset along with our code. A sample of object crops and their corresponding captions are shown in Figure \ref{fig:captions}.

\subsection{Filtering for Relevant Objects}
Indoor environments such as homes can consist of hundreds of objects, of which only a few are relevant to a given language query or task. Thus, inspired by the ability for humans to filter out irrelevant objects and the success of past works \cite{fang2024transcrib3d, hsu2023ns3d}, we implement an LLM-based filtering module consisting of two queries. Given an input command like ``The nightstand to the right of the bed", the first query extracts object nouns and modifiers (e.g. nightstand and bed), and the second query returns the relevant object instances from the list of IDs and names provided by the perception module (in this case returning all nightstands and beds in the scene). We use Mistral Large 2 \cite{mistralLargeEnough} for these steps, filtering objects based on their text descriptions, to best leverage the ability of LLMs to process textual information.

\subsection{Spatial Reasoning Toolbox}

With only the relevant objects extracted, the subset of objects and their captions are then fed to an LLM reasoning agent as a list of $n_o$ objects, where each object $o_i$ is represented by the list of attributes: $\{\texttt{id}, \texttt{name}, \texttt{caption}, c_x, c_y, c_z, \texttt{size}\}$, where \texttt{id} is a unique integer identifier for the object, $(c_x, c_y, c_z)$ are discretized coordinates of the object center, and \texttt{size} is the area of the largest face. 

However, directly prompting off-the-shelf LLMs for 3D referential grounding with this scene representation results in poor performance due to their limitations in spatial and mathematical reasoning: for example, when querying "the nightstand to the left of the bed", the LLM picks the one with the smallest $x$ value. As a result, we abstract spatial reasoning away from the LLM by creating a \textbf{spatial reasoning toolbox} consisting of heuristic search functions that find objects referred to by a set of elementary spatial relations. The key idea is that when referring to an object using relative spatial relations, only a fixed set of relations are needed. Thus, defining this set of heuristic functions is sufficient for handling diverse inter-object referential scenarios. The search functions and their arguments are listed in Table \ref{tab:spatial_relations}.

\begin{table}[htbp]
    \begin{tabularx}{\linewidth}{|X|}
        \hline
        \textbf{Spatial Search Functions} \\
        \hline
        find\_near(target, anchor)\textsuperscript{*}\\
        \hline
        find\_between(target, anchor1, anchor2)\textsuperscript{*} \\
        \hline
        find\_above(target, anchor)\textsuperscript{*} \\
        \hline
        find\_below(target, anchor)\textsuperscript{*} \\
        \hline
        order\_bottom\_to\_top(targets)\textsuperscript{*} \\
        \hline
        order\_smallest\_to\_largest(targets)\textsuperscript{*} \\
        \hline
        find\_objects\_near\_room\_corner(targets)\textsuperscript{*} \\
        \hline
        find\_left(target, anchor)\textsuperscript{\dag}  \\
        \hline
        find\_right(target, anchor)\textsuperscript{\dag}  \\
        \hline
        order\_left\_to\_right(target, anchor)\textsuperscript{\dag} \\
        \hline
    \end{tabularx}
    \caption{Heuristic search functions in the spatial toolbox. View-independent functions are marked with an asterisk (*), and view-dependent functions are marked with a dagger (\dag).}
    \label{tab:spatial_relations}
\end{table}

The LLM is prompted with an in-context example to decompose a referential statement into a series of search calls and choose a single object ID from the returned lists of IDs. For example, given the query ``Find the computer near the desk with a printer on it", the LLM first calls \verb|find_below(desk, printer)| returning \verb|[2]|, then \verb|find_near(computer, 2)| returning \verb|[3, 4]|, finally picking the computer with ID 3. We note a few implementation details regarding the search functions:

\begin{enumerate}
    \item For statements employing view-dependent relationships, like left and right, we tackle the case where no specific observer viewpoint is given (common in Nr3D), and assume the relationship is unambiguous from any feasible viewing direction. We therefore determinine left/right direction from the signed angle formed by the nearest point in free space and the target and anchor's centroids.
    \item We use the area of the largest face in \verb|order_smallest_to_largest|, which is more intuitive than volume for flat objects.
    \item All search functions have access to the full output of the perception module including object names and 3D bounding boxes, which is partially hidden from the LLM. This allows us to develop more complex logic for search functions without affecting LLM reasoning, which is an advantage of our approach.
\end{enumerate}
Our accompanying repository contains further implementation details on the spatial search functions.

\subsection{Parsing for Action Execution} \label{sec:codegen}
For our benchmark results, the LLM is prompted to only output a single object ID denoting the chosen object. For our real-world deployed pipeline, the LLM may output a series of either \verb|go_near| or \verb|go_between| function calls which take object IDs as arguments and generate waypoints in the scene which are sent to an obstacle-avoidant downstream planner for sequential navigation. \verb|go_near(target)| places a waypoint at the closest point in traversable space to the target's center, while \verb|go_between(target1, target2)| places one near the midpoint between the two targets.

\section{Experimental Setup}
We quantitatively evaluate our method on two 3D object-referential datasets, ReferIt3D \cite{achlioptas2020referit3d} and IRef-VLA \cite{zhang2025irefvlabenchmarkinteractivereferential}. Both datasets consist of utterances describing a target object in a ScanNet \cite{dai2017scannet} scene using spatial relations. In particular, ReferIt3D is split into Sr3D, which consists of synthetically generated utterances from five relation categories while Nr3D consists of natural language statements collected from humans, with unconstrained methods of describing target objects. Statements are categorized as ``Easy"/``Hard" based on the number of ``distractor" objects of the same class as the target object in the scene and also ``View-Dependent"/``View-Independent". IRef-VLA consists entirely of template-generated statements that are view-independent but contains utterances for a diverse set of 3D scans and enforces every statement to contain a spatial relation. The set of spatial relations is expanded to include eight total relations, including ternary relations (e.g. ``between") and numerical relations (e.g. ``second closest"). Utterances may also contain attributes such as color and size if needed to disambiguate the target object from distractors. Additionally, we deploy SORT3D on two ground vehicles, and validate the system's generalizability by testing navigation commands containing references to spatial relations, references to object attributes, and implicit or indirect requests in three previously unseen indoor environments.

\subsection{Referential Grounding on Benchmark Datasets}
We test our model on both ReferIt3D subsets and the subset of IRef-VLA using ScanNet scenes and compare to SOTA baselines. On each data subset, we evaluate our model on 200 sampled statements\footnote{We use a subset of the test set due to the cost of LLM evaluation on the full test set and the need for multiple runs to obtain variance statistics}, sampled to match the distribution of Easy, Hard, View-Dependent, and View-Independent statements in the original ReferIt3D test dataset. We focus our comparison against Transcrib3D \cite{fang2024transcrib3d} which, to the best of our knowledge, is the best performing model on ReferIt3D to date. For a fair comparison, we run Transcrib3D with the same two LLMs we use on the same test splits\footnote{the GPT4 model used in their work is now a legacy model. We run their method with GPT4o instead.}. For our methods, we conduct multiple trials on each data split to measure variance in LLMs, reported with standard deviation values on the grounding accuracy, which we note that other LLM-based methods do not report. 


\section{Results and Discussion}
\subsection{Referential Grounding on Benchmark Datasets}
The grounding accuracy on ReferIt3D is shown in Tables \ref{tab:Nr3D} and \ref{tab:Sr3D}, and accuracy on IRef-VLA is shown in Table \ref{tab:IRef-VLA}. We see that our method achieves higher accuracy with GPT-4o as the LLM backend and is on par with SOTA methods on View-Dependent statements in Nr3D and Hard statements in IRef-VLA while requiring no data to train. We also note that the use of LLMs introduces variance between trials, affecting grounding accuracy up to 6\%. 

While supervised baselines such as ViL3DRel \cite{chen2022language}, 3D-VisTA \cite{zhu20233d}, and SceneVerse \cite{jia2024sceneverse} report slightly higher overall grounding accuracies, these methods are explicitly trained on ReferIt3D data. Similarly, while Transcrib3D reports higher accuracies, it relies on guiding principles \cite{fang2024transcrib3d} that are tailored to the language used in Nr3D and Sr3D, which improve performance on those benchmarks.

\begin{figure}[htbp]
    \centering
    \includegraphics[width=\linewidth]{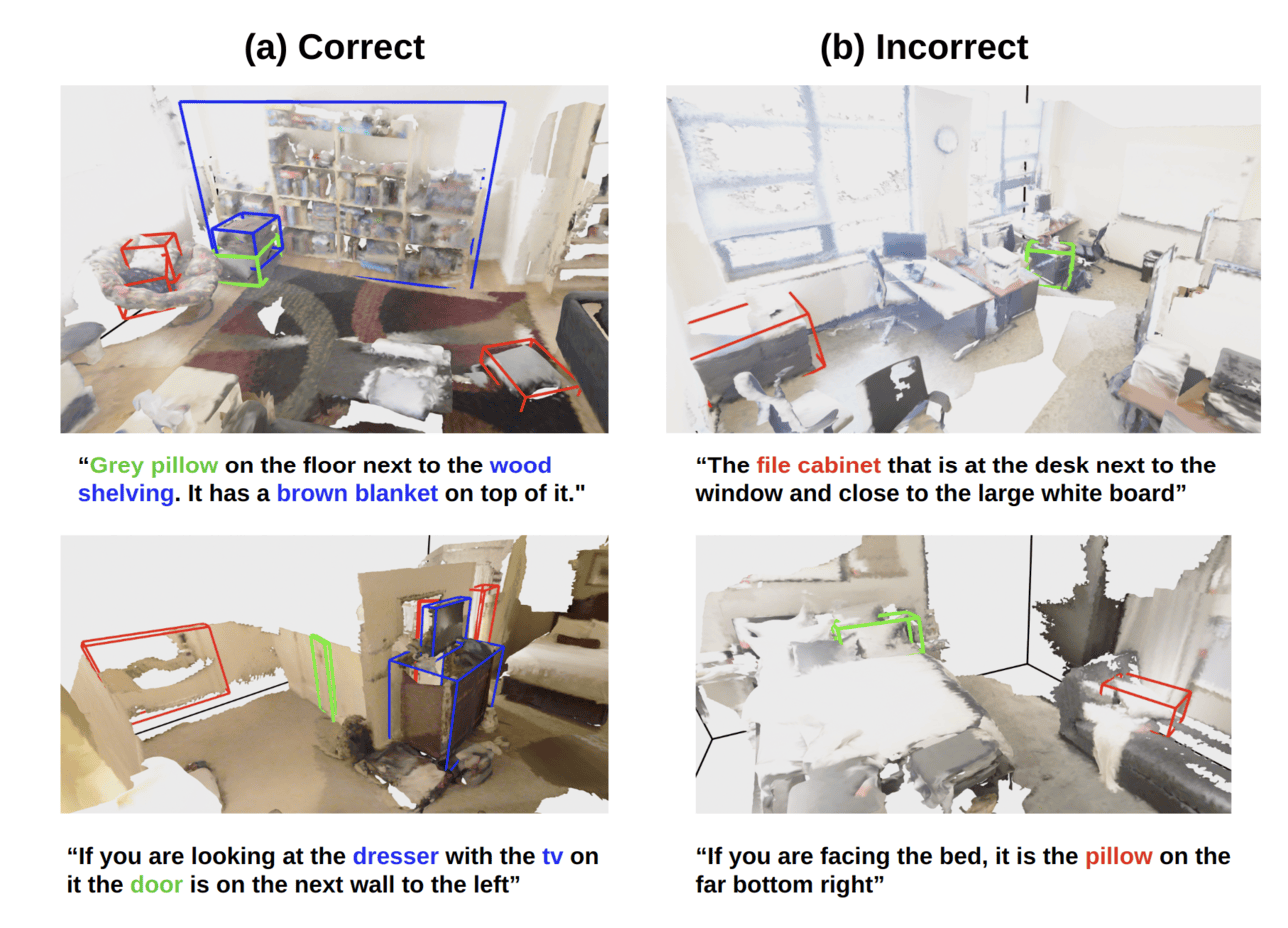}
    \caption{Correct (a) and incorrect (b) grounding examples. Top left and bottom left respectively show correctly grounded view-independent and view-dependent statements. Top right and bottom right are two examples of model logic failing: in the top right image, the model picks out the desk closest to a window, but not near the whiteboard. In the bottom right, the model fails at pragmatics, picking out the rightmost pillow, instead of recognizing that the sentence implies choosing a pillow on the bed.}
    \label{fig:correct-incorrect}
\end{figure}

In contrast, our approach is purely zero-shot, requiring only one single example of how to use the spatial reasoning toolbox, which does not have to be from a particular dataset, and we employed no dataset-specific training or fine-tuning. Despite this, by leveraging foundation models to obtain object semantic attributes and mapping spatial reasoning into sequential reasoning, our spatial reasoning toolbox approach achieves overall performance comparable to SOTA supervised and fine-tuned methods on Nr3D, and performance on view-dependent statements on par with Transcrib3D. On Sr3D, SORT3D surpasses SOTA supervised training methods and achieves close overall performance to Transcrib3D while surpassing it in view-dependent accuracy. This demonstrates the effectiveness of our approach at handling spatial reasoning where viewpoint anchoring is required. We see that SORT3D is able to explainably resolve complex view-dependent relations with multiple anchors and complex semantic descriptions (Figure \ref{fig:correct-incorrect}-a), while also providing explainable model failure points by analyzing its chain of thought reasoning (Figure \ref{fig:correct-incorrect}-b).

On IRef-VLA, our method surpasses other methods on Hard statements by a large margin. IRef-VLA contains a large set of statements using size and color descriptions when referring to objects, which our method effectively grounds by utilizing object filtering, semantic attributes captured by captions, and the spatial reasoning toolbox to identify target objects with multiple distractors in the scene. Transcrib3D, when used in its zero-shot formulation with principles and not fine-tuned on the dataset's hard statements  \cite{fang2024transcrib3d}, fails to generalize well despite using the same LLM backbone. The results on IRef-VLA strongly support the benefits of incorporating 2D captions, as they provide critical, fine-grained object-level attributes that refine the referential grounding.

\begin{table}[htbp]
    \centering
    \small
    \caption{Performance on the Nr3D dataset from the ReferIt3D benchmark. Asterisks (*) indicate results reported from the paper directly. ``View Dep." and ``View Ind." stand for view-dependent and view-independent respectively.} 
    \begin{tabularx}{\linewidth}{p{3.2cm}ccc}
        \toprule
        \multicolumn{4}{c}{Nr3D} \\
        \midrule
        Method & Overall &  \makecell{View Dep.} & \makecell{View Ind.} \\
        \midrule
        \textbf{Supervised Methods} \\
        NS3D* \cite{hsu2023ns3d} & 62.7 & 62.0 & - \\
        ViL3DRel*~\cite{chen2022language} & 64.4 & 62.0 & 64.5 \\ 
        3D-VisTA*~\cite{zhu20233d} & 64.2 & 61.5 & 65.1 \\
        SceneVerse-GPS* \cite{jia2024sceneverse} & 64.9 & 56.9 & 67.9 \\ 
        \midrule
        \textbf{Zero-Shot Methods} \\
        ZSVG3D* \cite{yuan2024visual} & 39.0 & 36.8 & 40.0 \\
        VLM-Grounder* \cite{xu2024vlm} & 48.0 & 45.8 &  49.4 \\
        CSVG* \cite{yuan2024solving} & 59.2 & 53.0 & 62.5 \\
        Transcrib3D* \cite{fang2024transcrib3d} (GPT-4) & \textbf{70.2} & \textbf{60.1} & \textbf{75.4} \\
        \midrule
        Transcrib3D \scriptsize{(GPT-4o)}& \textbf{65.6} & \textbf{63.3} & \textbf{66.7} \\
        Transcrib3D \scriptsize{(Mistral)} & \underline{63.8} & 57.1 & \textbf{66.7} \\
        \textbf{Ours} \textbf{\scriptsize{(GPT-4o)}} & 62.0$\pm$1.2 & 56.6$\pm$0.0 & \underline{64.3$\pm$1.7} \\
        \textbf{Ours} \textbf{\scriptsize{(Mistral)}} & 61.6$\pm$0.3 & \underline{59.4$\pm$0.9} & 62.6$\pm$0.9 \\
        \bottomrule
    \end{tabularx}
    \label{tab:Nr3D}
\end{table}

\begin{table}[htbp]
    \centering
    \small
    \caption{Performance on the Sr3D dataset from the ReferIt3D benchmark. Asterisks (*) indicate results reported from the paper directly.} 
    \begin{tabularx}{\linewidth}{p{3.2cm}ccc}
        \toprule
        \multicolumn{4}{c}{Sr3D} \\
        \midrule
        Method & Overall &  \makecell{View Dep.} & \makecell{View Ind.} \\
        \midrule
        \textbf{Supervised Methods} \\
        ViL3DRel*~\cite{chen2022language} & 72.8 & 63.8 & 73.2 \\ 
        3D-VisTA*~\cite{zhu20233d} & 76.4 & 58.9 & 77.3 \\
        SceneVerse-GPS* \cite{jia2024sceneverse} & 77.5 & 62.8 & 78.2 \\   
        \midrule
        \textbf{Zero-Shot Methods} \\
        Transcrib3D* \cite{fang2024transcrib3d} \scriptsize{(GPT-4)} & 98.4 & 98.2 & 98.4\\
        \midrule
         Transcrib3D \scriptsize{(GPT-4o)}& \textbf{96.5} &  88.9 & \textbf{96.9} \\
         Transcrib3D \scriptsize{(Mistral)} & \underline{96.0} & 77.8 & \textbf{96.9} \\
         \textbf{Ours \scriptsize{(Mistral)}} & 92.0$\pm$0.7 &  \underline{90.9$\pm$0.0} & \underline{92.2$\pm$0.8}\\
         \textbf{Ours \scriptsize{(GPT-4o)}} & 92.0$\pm$0.0 & \textbf{95.5$\pm$0.0} & 91.6$\pm$0.0 \\
        \bottomrule
    \end{tabularx}  
    \label{tab:Sr3D}
\end{table}

\begin{table}[!ht]
    \centering
    \small
    \caption{Grounding accuracy on IRef-VLA test subset} 
    \begin{tabularx}{\linewidth}{p{3.2cm}ccc}
        \toprule
        \multicolumn{4}{c}{IRef-VLA} \\
        \midrule
        Method & Overall &  \makecell{Easy} & \makecell{Hard} \\
        \midrule
        Transcrib3D \scriptsize{(Mistral)} & 70.5 & \underline{76.25} & 47.5 \\
        Transcrib3D \scriptsize{(GPT-4o)} & \textbf{77.5} & \textbf{82.5} & 57.5 \\
        Ours \scriptsize{(Mistral)} & 69.0$\pm$0.7 & 70.0$\pm$0.8 & \underline{65.0$\pm$0.0} \\
        Ours \scriptsize{(GPT-4o)} & \underline{71.8$\pm$1.8} & 71.0$\pm$1.3 & \textbf{75.0}$\pm$\textbf{3.5} \\
        \bottomrule
    \end{tabularx}  
    \label{tab:IRef-VLA}
\end{table}


\subsection{Ablation of Captioning Module}
We evaluate the effect on grounding accuracy of adding open-vocabulary captions generated from 2D images of objects in the scene. We augment the Transcrib3D \cite{fang2024transcrib3d} baseline model with our captions for each object as additional information passed into the LLM reasoner. We hypothesize that these finer descriptions of object attributes will help the model in disambiguating objects when given free-form referential statements. We test this hypothesis through an ablation study using GPT-4o with both our method and Transcrib3D on Nr3D, shown in Table \ref{tab:captions-ablation}. In both Transcrib3D and our method, we observe consistent and significant improvements across all statements types. For our approach specifically, the addition of captions improves performance the most (11\%) on view-dependent statements. These results demonstrate that understanding detailed object attributes is important for effective 3D grounding and leveraging 2D VLMs is an effective method for this. Many referential statements rely on subtle distinctions—such as color, shape, texture, or affordance—that traditional 3D models often miss.

\begin{table*}[htbp]
    \centering
    \small
    \caption{Grounding accuracy with and without captions on Nr3D} 
    \begin{tabularx}{0.9\linewidth}{lcccccccc}
        \toprule
        & \multicolumn{5}{c}{Nr3D} \\
        \cmidrule(lr){2-6}
        Method & Overall & Easy & Hard & \makecell{View Dep.} & \makecell{View Ind.} \\
        \midrule
         Transcrib3D \scriptsize{(GPT-4o)}& 58.5 & 67.5 & 45.0 & 54.0 & 60.6 \\
         Transcrib3D \scriptsize{(GPT-4o + captions)}& 61.0 ($\uparrow$ 4.3\%) & 70.8 ($\uparrow$ 4.9\%) & 46.3 ($\uparrow$ 2.9\%) & 55.4 ($\uparrow$ 2.6\%) & 63.5 ($\uparrow$ 4.8\%) \\
        \midrule
         Ours \scriptsize{(GPT-4o)} & 54.5 & 59.2 & 47.5 & 50.7 & 56.2 \\
         Ours \scriptsize{(GPT-4o + captions)} & 60.5 ($\uparrow$ 11.0\%) & 64.2 ($\uparrow$ 8.4\%) & 55.0 ($\uparrow$ 5.3\%) & 56.6 ($\uparrow$ \textbf{11.6\%}) & 62.3 ($\uparrow$ 10.9\%) \\
        \bottomrule
    \end{tabularx}  
    \label{tab:captions-ablation}
\end{table*}

\subsection{Real-World Validation} \label{sec:real-world}

\begin{figure}[htbp]
    \centering
    \includegraphics[width=\linewidth]{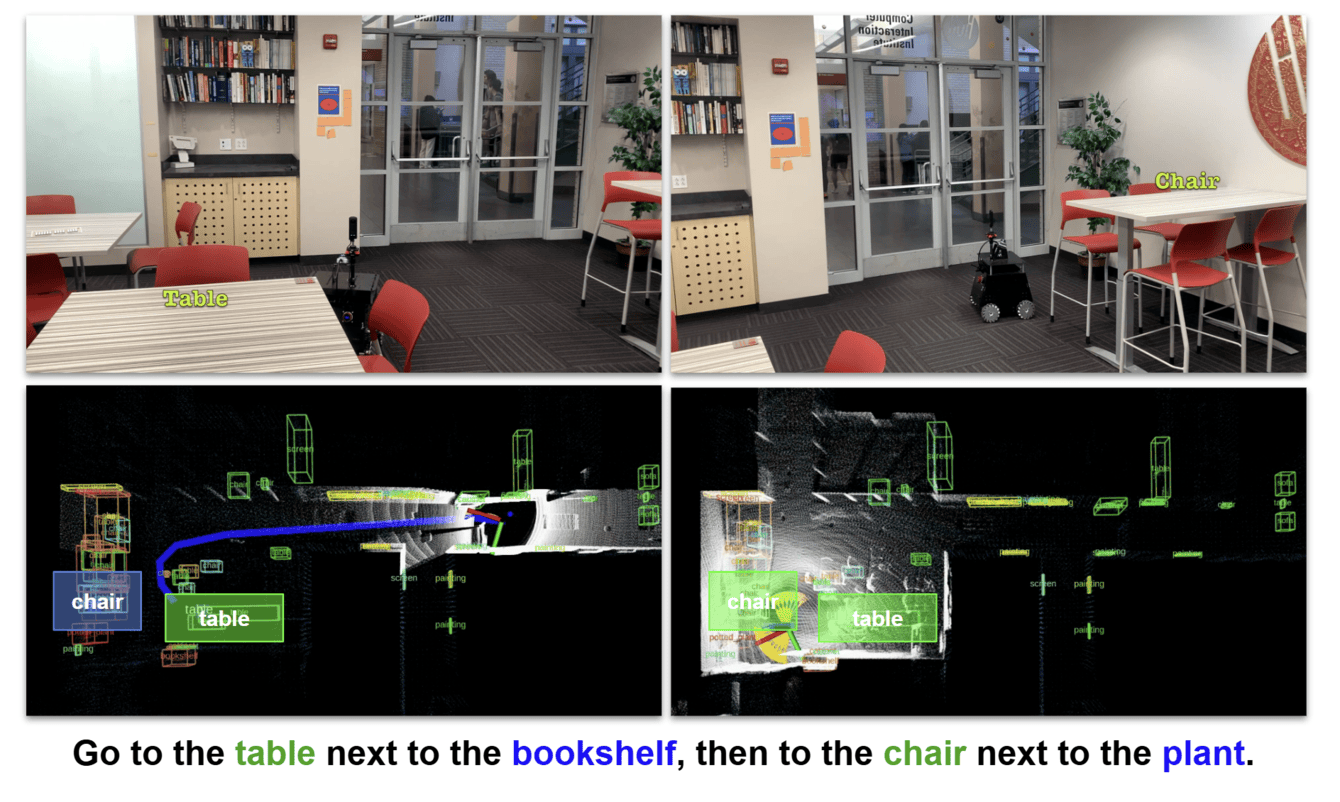}
    \caption{Navigation on the mecanum robot in the university corridor given the statement ``Go to the table next to the bookshelf, then to the chair next to the plant." The system successfully navigates to both spatially referenced objects.}
    \label{fig:corridor_table_bookshelf}
\end{figure}

\begin{figure}[htbp]
    \centering
    \includegraphics[width=\linewidth]{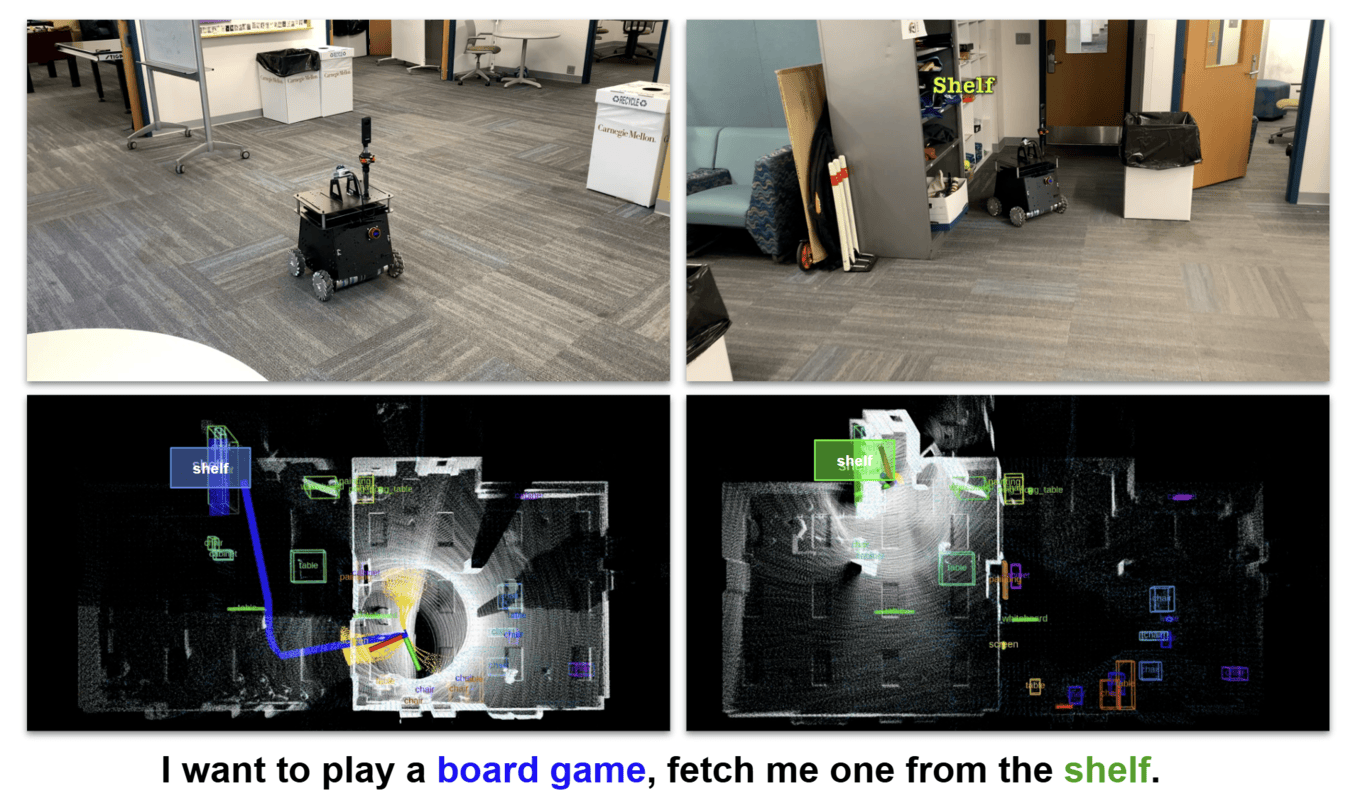}
    \caption{Navigation on the mecanum robot in the student lounge given the statement ``I want to play a board game, fetch me one from the shelf." The system successfully navigates to the shelf with the board game on it by referring to semantic caption descriptions.}
    \label{fig:robolounge_boardgame_shelf}
\end{figure}

To validate SORT3D's pipeline in the real-world, we implement the full system on two autonomous mobile robots: a mecanum-wheeled robot and a differential drive robot with a wheelchair base, each equipped with 360-degree cameras, respective Livox and Velodyne LiDARS, onboard Intel NUCs for low level autonomy, and RTX 4090s with 16GB and 24GB of VRAM respectively. We validate our system in two previously unseen indoor environments: a student lounge (Figure \ref{fig:robolounge_boardgame_shelf}, and a university corridor (Figure \ref{fig:corridor_table_bookshelf}). We attempt two different types of navigational queries, which target the system's ability to ground statements that employ both spatial references and object semantic attributes. Before issuing a query, we navigate each robot around the scene to build a semantic map after prompting the open vocabulary detector with the names and synonyms of objects in the scene (shown in RViz in the bottom of Figures \ref{fig:robolounge_boardgame_shelf} and \ref{fig:corridor_table_bookshelf}), and collect image crops for each detected object. As an implementation detail, captions are batch-generated only after a user query is typed into the system to speed up semantic mapping and decrease overall power consumption. RViz visualizations of the instance level semantic maps, the objects and corresponding waypoints chosen by the grounding model for each statement, and pictures of each platform navigating through its environment are shown in Figures \ref{fig:robolounge_boardgame_shelf} and \ref{fig:corridor_table_bookshelf}. In each statement we test, SORT3D successfully grounds one or more referenced objects, demonstrating the versatility of our approach for grounding complex expressions involving spatial references and semantic attributes in previously unseen scenes. Further experiments on more environments and types of statements are found in our accompanying repository and videos.

\section{Limitations and Future Work}
We acknowledge that limitations exist within our approach. First, our system relies on internet access to call online APIs, which is a reasonable assumption for indoor environments like homes and offices, but may not hold for other environments where generalizable robot systems may need to be deployed. To address this limitation, the modular design of our pipeline allows for the LLMs to be easily replaced by smaller local models.

We also recognize that our evaluation set is limited. There are few existing benchmarks designed to rigorously test online 3D referential grounding with diverse, attribute-rich natural language. This constraint makes it challenging to fully assess the generalizability of our approach across different environments or while deployed on a system. Evaluating our method in a simulated environment with real-time interactions would provide deeper insights into its effectiveness and adaptability as a system. It would also allow us to investigate more practical settings such as multi-turn grounding scenarios and failure correction.

\section{Conclusion}

We introduce SORT3D, a robust, data-efficient, and deployable method for 3D referential grounding with complex view-dependent reasoning. Our framework combines 2D visual features, LLM-based filtering, and a heuristics-driven spatial reasoning toolbox to enhance LLMs’ ability to handle spatial and mathematical reasoning in 3D scenes. SORT3D achieves results that are competitive to SOTA methods on view-dependent and attribute-based statements while outperforming other fully zero-shot methods. We further demonstrate its performance and generalization  by deploying the pipeline on a robotic system for real-time indoor navigation.
\FloatBarrier



\bibliographystyle{IEEEtran}
\bibliography{refs}

\begin{thebibliography}{10}
\providecommand{\url}[1]{#1}
\csname url@rmstyle\endcsname
\providecommand{\newblock}{\relax}
\providecommand{\bibinfo}[2]{#2}
\providecommand\BIBentrySTDinterwordspacing{\spaceskip=0pt\relax}
\providecommand\BIBentryALTinterwordstretchfactor{4}
\providecommand\BIBentryALTinterwordspacing{\spaceskip=\fontdimen2\font plus
\BIBentryALTinterwordstretchfactor\fontdimen3\font minus \fontdimen4\font\relax}
\providecommand\BIBforeignlanguage[2]{{%
\expandafter\ifx\csname l@#1\endcsname\relax
\typeout{** WARNING: IEEEtran.bst: No hyphenation pattern has been}%
\typeout{** loaded for the language `#1'. Using the pattern for}%
\typeout{** the default language instead.}%
\else
\language=\csname l@#1\endcsname
\fi
#2}}

\bibitem{achlioptas2020referit3d}
P.~Achlioptas, A.~Abdelreheem, F.~Xia, M.~Elhoseiny, and L.~Guibas, ``Referit3d: Neural listeners for fine-grained 3d object identification in real-world scenes,'' in \emph{Computer Vision--ECCV 2020: 16th European Conference, Glasgow, UK, August 23--28, 2020, Proceedings, Part I 16}.\hskip 1em plus 0.5em minus 0.4em\relax Springer, 2020, pp. 422--440.

\bibitem{ramakrishnan2021habitat}
S.~K. Ramakrishnan, A.~Gokaslan, E.~Wijmans, O.~Maksymets, A.~Clegg, J.~Turner, E.~Undersander, W.~Galuba, A.~Westbury, A.~X. Chang, \emph{et~al.}, ``Habitat-matterport 3d dataset (hm3d): 1000 large-scale 3d environments for embodied ai,'' \emph{arXiv preprint arXiv:2109.08238}, 2021.

\bibitem{jain2022bottom}
A.~Jain, N.~Gkanatsios, I.~Mediratta, and K.~Fragkiadaki, ``Bottom up top down detection transformers for language grounding in images and point clouds,'' in \emph{European Conference on Computer Vision}.\hskip 1em plus 0.5em minus 0.4em\relax Springer, 2022, pp. 417--433.

\bibitem{huang2022multi}
S.~Huang, Y.~Chen, J.~Jia, and L.~Wang, ``Multi-view transformer for 3d visual grounding,'' in \emph{Proceedings of the IEEE/CVF Conference on Computer Vision and Pattern Recognition}, 2022, pp. 15\,524--15\,533.

\bibitem{zhu20233d}
Z.~Zhu, X.~Ma, Y.~Chen, Z.~Deng, S.~Huang, and Q.~Li, ``3d-vista: Pre-trained transformer for 3d vision and text alignment,'' in \emph{Proceedings of the IEEE/CVF International Conference on Computer Vision}, 2023, pp. 2911--2921.

\bibitem{fang2024transcrib3d}
J.~Fang, X.~Tan, S.~Lin, I.~Vasiljevic, V.~Guizilini, H.~Mei, R.~Ambrus, G.~Shakhnarovich, and M.~R. Walter, ``Transcrib3d: 3d referring expression resolution through large language models,'' \emph{arXiv preprint arXiv:2404.19221}, 2024.

\bibitem{hsu2023ns3d}
J.~Hsu, J.~Mao, and J.~Wu, ``Ns3d: Neuro-symbolic grounding of 3d objects and relations,'' in \emph{Proceedings of the IEEE/CVF Conference on Computer Vision and Pattern Recognition}, 2023, pp. 2614--2623.

\bibitem{chen2024grounded}
Y.~Chen, S.~Yang, H.~Huang, T.~Wang, R.~Xu, R.~Lyu, D.~Lin, and J.~Pang, ``Grounded 3d-llm with referent tokens,'' \emph{arXiv preprint arXiv:2405.10370}, 2024.

\bibitem{yuan2024solving}
Q.~Yuan, J.~Zhang, K.~Li, and R.~Stiefelhagen, ``Solving zero-shot 3d visual grounding as constraint satisfaction problems,'' \emph{arXiv preprint arXiv:2411.14594}, 2024.

\bibitem{yang2024llm}
J.~Yang, X.~Chen, S.~Qian, N.~Madaan, M.~Iyengar, D.~F. Fouhey, and J.~Chai, ``Llm-grounder: Open-vocabulary 3d visual grounding with large language model as an agent,'' in \emph{2024 IEEE International Conference on Robotics and Automation (ICRA)}.\hskip 1em plus 0.5em minus 0.4em\relax IEEE, 2024, pp. 7694--7701.

\bibitem{xu2024vlm}
R.~Xu, Z.~Huang, T.~Wang, Y.~Chen, J.~Pang, and D.~Lin, ``Vlm-grounder: A vlm agent for zero-shot 3d visual grounding,'' \emph{arXiv preprint arXiv:2410.13860}, 2024.

\bibitem{li2024seeground}
R.~Li, S.~Li, L.~Kong, X.~Yang, and J.~Liang, ``Seeground: See and ground for zero-shot open-vocabulary 3d visual grounding,'' \emph{arXiv preprint arXiv:2412.04383}, 2024.

\bibitem{dai2017scannet}
A.~Dai, A.~X. Chang, M.~Savva, M.~Halber, T.~Funkhouser, and M.~Nie{\ss}ner, ``Scannet: Richly-annotated 3d reconstructions of indoor scenes,'' in \emph{Proceedings of the IEEE conference on computer vision and pattern recognition}, 2017, pp. 5828--5839.

\bibitem{Matterport3D}
A.~Chang, A.~Dai, T.~Funkhouser, M.~Halber, M.~Niessner, M.~Savva, S.~Song, A.~Zeng, and Y.~Zhang, ``Matterport3d: Learning from {RGB-D} data in indoor environments,'' \emph{International Conference on 3D Vision (3DV)}, 2017.

\bibitem{batra2020objectnav}
D.~Batra, A.~Gokaslan, A.~Kembhavi, O.~Maksymets, R.~Mottaghi, M.~Savva, A.~Toshev, and E.~Wijmans, ``Objectnav revisited: On evaluation of embodied agents navigating to objects,'' \emph{arXiv preprint arXiv:2006.13171}, 2020.

\bibitem{wei2022chain}
J.~Wei, X.~Wang, D.~Schuurmans, M.~Bosma, F.~Xia, E.~Chi, Q.~V. Le, D.~Zhou, \emph{et~al.}, ``Chain-of-thought prompting elicits reasoning in large language models,'' \emph{Advances in neural information processing systems}, vol.~35, pp. 24\,824--24\,837, 2022.

\bibitem{zhang2025irefvlabenchmarkinteractivereferential}
\BIBentryALTinterwordspacing
H.~Zhang, N.~Zantout, P.~Kachana, J.~Zhang, and W.~Wang, ``Iref-vla: A benchmark for interactive referential grounding with imperfect language in 3d scenes,'' 2025. [Online]. Available: \url{https://arxiv.org/abs/2503.17406}
\BIBentrySTDinterwordspacing

\bibitem{chen2020scanrefer}
D.~Z. Chen, A.~X. Chang, and M.~Nie{\ss}ner, ``Scanrefer: 3d object localization in rgb-d scans using natural language,'' in \emph{European conference on computer vision}.\hskip 1em plus 0.5em minus 0.4em\relax Springer, 2020, pp. 202--221.

\bibitem{jia2024sceneverse}
B.~Jia, Y.~Chen, H.~Yu, Y.~Wang, X.~Niu, T.~Liu, Q.~Li, and S.~Huang, ``Sceneverse: Scaling 3d vision-language learning for grounded scene understanding,'' \emph{arXiv preprint arXiv:2401.09340}, 2024.

\bibitem{chen2022language}
S.~Chen, P.-L. Guhur, M.~Tapaswi, C.~Schmid, and I.~Laptev, ``Language conditioned spatial relation reasoning for 3d object grounding,'' \emph{Advances in neural information processing systems}, vol.~35, pp. 20\,522--20\,535, 2022.

\bibitem{yuan2024visual}
Z.~Yuan, J.~Ren, C.-M. Feng, H.~Zhao, S.~Cui, and Z.~Li, ``Visual programming for zero-shot open-vocabulary 3d visual grounding,'' in \emph{Proceedings of the IEEE/CVF Conference on Computer Vision and Pattern Recognition}, 2024, pp. 20\,623--20\,633.

\bibitem{3dllm}
\BIBentryALTinterwordspacing
Y.~Hong, H.~Zhen, P.~Chen, S.~Zheng, Y.~Du, Z.~Chen, and C.~Gan, ``3d-llm: Injecting the 3d world into large language models,'' 2023. [Online]. Available: \url{https://arxiv.org/abs/2307.12981}
\BIBentrySTDinterwordspacing

\bibitem{zhou2024navgpt}
G.~Zhou, Y.~Hong, and Q.~Wu, ``Navgpt: Explicit reasoning in vision-and-language navigation with large language models,'' in \emph{Proceedings of the AAAI Conference on Artificial Intelligence}, vol.~38, no.~7, 2024, pp. 7641--7649.

\bibitem{lin2024navcot}
B.~Lin, Y.~Nie, Z.~Wei, J.~Chen, S.~Ma, J.~Han, H.~Xu, X.~Chang, and X.~Liang, ``Navcot: Boosting llm-based vision-and-language navigation via learning disentangled reasoning,'' \emph{arXiv preprint arXiv:2403.07376}, 2024.

\bibitem{liu2024groundingdinomarryingdino}
\BIBentryALTinterwordspacing
S.~Liu, Z.~Zeng, T.~Ren, F.~Li, H.~Zhang, J.~Yang, Q.~Jiang, C.~Li, J.~Yang, H.~Su, J.~Zhu, and L.~Zhang, ``Grounding dino: Marrying dino with grounded pre-training for open-set object detection,'' 2024. [Online]. Available: \url{https://arxiv.org/abs/2303.05499}
\BIBentrySTDinterwordspacing

\bibitem{ravi2024sam2segmentimages}
\BIBentryALTinterwordspacing
N.~Ravi, V.~Gabeur, Y.-T. Hu, R.~Hu, C.~Ryali, T.~Ma, H.~Khedr, R.~Rädle, C.~Rolland, L.~Gustafson, E.~Mintun, J.~Pan, K.~V. Alwala, N.~Carion, C.-Y. Wu, R.~Girshick, P.~Dollár, and C.~Feichtenhofer, ``Sam 2: Segment anything in images and videos,'' 2024. [Online]. Available: \url{https://arxiv.org/abs/2408.00714}
\BIBentrySTDinterwordspacing

\bibitem{yang2024qwen2technicalreport}
\BIBentryALTinterwordspacing
Q.~Team, ``Qwen2 technical report,'' 2024. [Online]. Available: \url{https://arxiv.org/abs/2407.10671}
\BIBentrySTDinterwordspacing

\bibitem{mistralLargeEnough}
M.~A. Team, ``Mistral large 2: The new generation of flag- ship model,'' \url{https://mistral.ai/news/mistral-large-2407/}, 2024, [Accessed 01-03-2025].

\end{thebibliography}

\end{document}